# SPECTRUM HOLE PREDICTION BASED ON HISTORICAL DATA: A NEURAL NETWORK APPROACH

Bara'u Gafai Najashi[1*], Feng Wenjiang[2] Mohammed Dikko Almustapha[3]

[1,2]College of Communication Engineering, Chongqing University,

Chongqing, China, Postal Code 400044

174 Shazheng Street, Shapingba District, Chongqing

[3]Department of Electrical and Computer Engineering, Ahmadu Bello University Zaria, Nigeria

## ABSTRACT

The concept of cognitive radio pioneered by Mitola promises to change the future of wireless communication especially in the area of spectrum management. Currently, the command and control strategy employed in spectrum assignment is too rigid and needs to be reviewed. Recent studies have shown that assigned spectrum is underutilized spectrally and temporally. Cognitive radio provides a viable solution whereby licensed users can share the spectrum with unlicensed users opportunistically without causing interference. Unlicensed users must be able to sense weather the channel is busy or idle, failure to do so will lead to interference to the licensed user. In this paper, a neural network based prediction model for predicting the channel status using historical data obtained during a spectrum occupancy measurement is presented. Genetic algorithm is combined with LM BP for increasing the probability of obtaining the best weights thus optimizing the network. The results obtained indicate high prediction accuracy over all bands considered.

## 1. INTRODUCTION

The field of wireless communication has witnessed tremendous evolution over the years. The demand for bandwidth has grown with the introduction of several wireless standards. This advancement has brought about a perceived spectrum scarcity. Recent studies have shown that there is ample spectrum available that is not being utilized. So the problem can be said to be spectrum under utilization and not spectrum scarcity. Cognitive radio is seen as a great contender for better spectrum utilization because of its ability to make spectrum sharing possible. A situation whereby a licensed user is allowed to share the spectrum with an unlicensed user also called a secondary user opportunistically without causing interference will surely increase the spectrum utilization. In order to achieve this, the unlicensed user must be able to sense when a channel is idle so that it can use, or busy so that it can vacate it. It has therefore become imperative to develop models capable of predicting the channel state accurately.

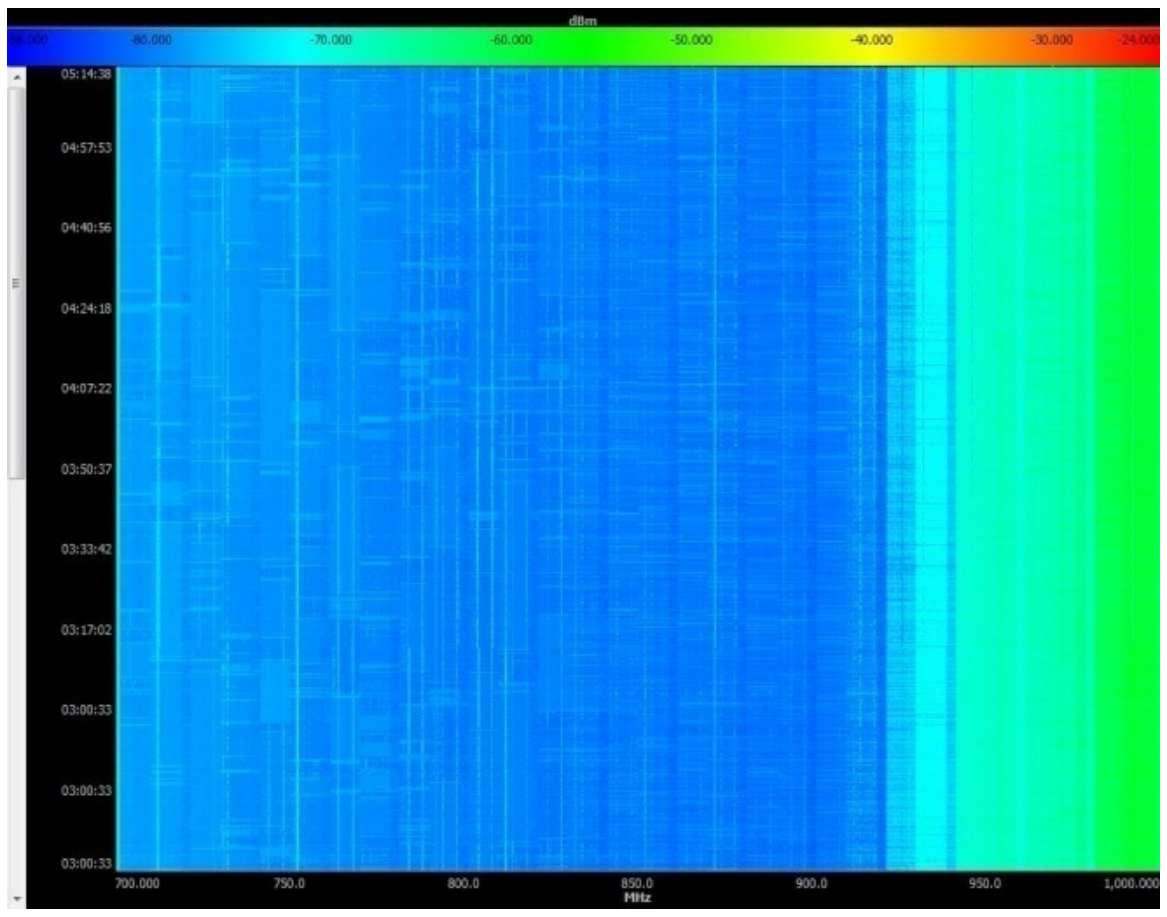

**Figure 1**: Waterfall figure showing sparsely used spectrum with low utilization level

Several spectrum prediction models have been proposed. Most of these works employ Markov chains for the prediction problem and assume that the primary user traffic follows a Poisson process [1][2][3]. In [4], a neural network based spectrum prediction using Multilayer Perceptron MLP was proposed. An hour long data was divided into 60 slots and converted into a time series. This was used to train and test the network. A practical spectrum behavior learning method based on MLP artificial neural network (ANN) was introduced in [5]. Performances were evaluated with an existing 7-days spectrum data set from a previous measurement, which was conducted in a metro city located in south

China. In this paper, a neural network based prediction model is presented. The training and test data were obtained from [10]. Unlike other neural network based prediction methods, a 12-hour long data was used for this work. The problem of weight selection which is common in neural networks was tackled using GA.

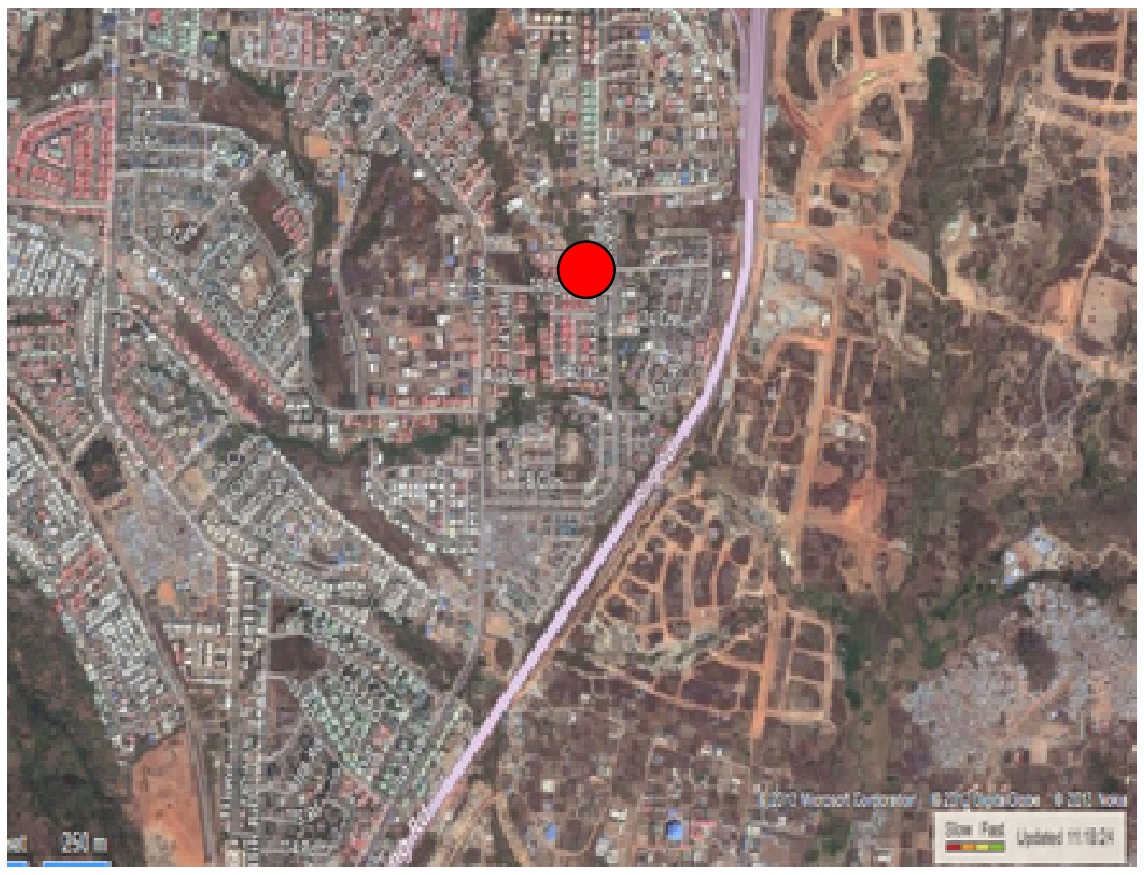

**Figure 2:** The red dot shows a bird's eye view of Gwarinpa, the measurement location in [10]

The rest of the paper is organized as follows. Section 2 provides a review of neural network, Multi-layer Perceptron and application of genetic algorithm in optimizing the interconnecting weights. In Section 3, the model used in this work is presented. Section 4 contains results obtained; conclusions are drawn from section 5.

## 2. NEURAL NETWORK

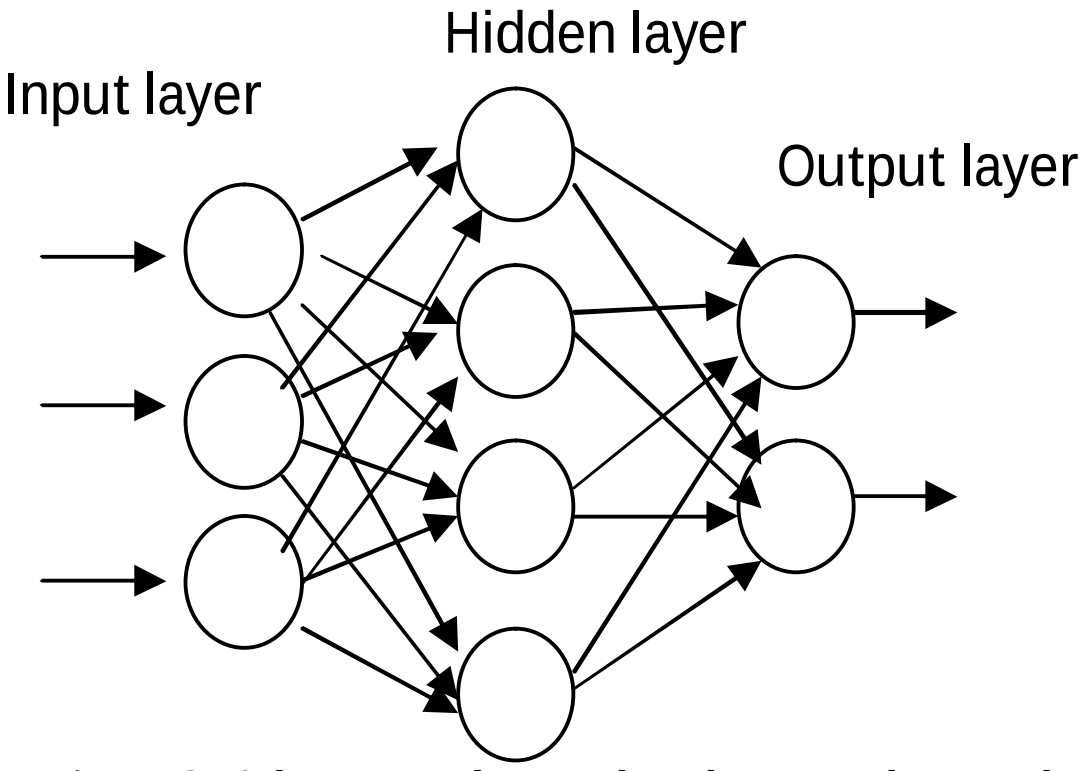

**Figure 3**: A basic neural network architecture showing the input, hidden and output layers

Artificial Neural Networks (ANNs) are non-linear mapping structures based on the function of the human brain. They are powerful tools for modeling and are widely used in prediction problems due to their simplicity in terms of training. Whereas other prediction schemes require continuous training, neural network are trained once in an offline fashion when the observed process is stationary [6]. The key element of this paradigm is the novel structure of the information processing system. It is composed of a large number of highly interconnected processing elements (neurons) working in unison to solve specific problems. An ANN is configured for a specific application, through a learning process. Learning in biological system involves adjustment to the synaptic connections that exist between the neurons [7]. There are basically 2 types of architectures namely, the feed forward and the recurrent networks. The feed forward architecture is where the connections are feed forward this means that a neuron will not accept an input from a neuron it has previously feed. The recurrent networks are those that allow feedback between neurons. For this work the feed forward architecture is used.

### 2.1 Multilayered Perceptron (MLP)

An MLP network is a multi layered structure consisting of an input, output and a few hidden layers. Each layer (excluding the input layer) consists of a number of computing units called neurons which calculate a weighted sum of the inputs and perform a nonlinear transformation on the sum. . The nonlinear transform is implemented using sigmoid functions (e.g. a hyperbolic tangent function). Neurons belonging to different layers are connected through adaptive weights. The output of a neuron j in the $i^{th}$ layer, denoted by $y^i_j$, can be represented as:

$$y_j^i = \frac{1 - \exp(-v_j^i)}{1 + \exp(-v_j^i)} \qquad (1)$$

Where $v_j^i = \sum_i y_i^{l-1} w_{ji}^l$ is the weighted sum of the inputs coming from the output of the neurons in the $(l-1)^{th}$ layer using the adaptive weights (or parameters) $w_{ji}^l$ connecting the neuron $j$ in the $l^{th}$ layer and neuron $i$ in the $(l-1)^{th}$ layer. Due to the nonlinear transform using the hyperbolic tangent function on $v_j^l$ in (1), $y_j^l$ lies in the range of [−1, +1]. If the inputs come from the input layer, $v_j^l$ is calculated using the corresponding inputs instead of $v_i^{l-1}$.The total number of inputs in the input layer is referred to as the order of the MLP network and is denoted by $\tau$. The number of hidden layers and the number of neurons in each layer depend on the application [8].

## 2.2 MLP Training

Once the architecture of a neural network has been determined for a particular application, the network is then ready for training. There are two types of training namely supervised and unsupervised training. In supervised learning, the input and output data is provided while in unsupervised learning the network has to make its decision on the input without any provision of the output. Over the years, MLP also known as a feed forward neural network has been applied successfully in several combinatorial problems. It works extremely well as a universal approximation in which input signal propagates in forward direction [11].

The general approach in learning involves presenting an input training pattern to an untrained network which is then used to get the output. The error is some scalar function of the weights that is minimized when the network outputs match the desired outputs. Since the actual output is known beforehand, the weights are adjusted to minimize the error so that the desired output could be reached. A Constant number 1/2 is added to do the mathematical derivation conveniently [5].

$$J(w) = \frac{1}{2}\sum_{k=1}^{c}(t_k - z_k)^2 \equiv \frac{1}{2}(t-z)^2 \qquad (2)$$

Where $t_k$is the desired output and $z_k$ is the actual output, t and z are the target network and output vectors with length c respectively. W represents the weights in the network. The initial weights are normally selected randomly and are updated in a way that reduces the error. This method is based on the gradient descent algorithm.

$$\vartriangle w = -\eta \frac{\partial J}{\partial w} \text{ or } (\vartriangle w_{pq} = -\eta \frac{\partial J}{\partial w_{pq}}) \qquad (3)$$

$\eta$ is the learning rate, which indicates the size of the change in weights. This iterative algorithm requires taking a weight vector at iteration m and updating it as

$$w(m+1) = w(m) + \vartriangle w(m) \qquad (4)$$

Where m indexes the particular pattern presentation. Considering the hidden-output weights $w_{kj}$ and applying chain rule differentiation

$$\frac{\partial J}{\partial w_{kj}} = \frac{\partial J}{\partial net_k}\frac{\partial net_k}{\partial w_{kj}} = \partial_k \frac{\partial net_k}{\partial w_{kj}} \qquad (5)$$

Where sensitivity of $z_k$ can be defined to be

$$\partial_k \equiv -\frac{\partial J}{\partial net_k} = -\frac{\partial J}{\partial z_k}\frac{\partial z_k}{\partial net_k} = (t_k - z_k)\, f'(net_k) \qquad (6)$$

Therefore, the weight update (learning rule) for the hidden-output weights given as

$$\vartriangle w_{kj} = \eta \partial_k y_j = \eta (t_k - z_k)\, f'(net_k) y_j \qquad (7)$$

Similarly, considering the input-hidden weights $w_{ji}$

$$\frac{\partial J}{\partial w_{ji}} = \frac{\partial J}{\partial y_j}\frac{\partial y_j}{\partial net_j}\frac{\partial net_j}{\partial w_{ji}} \qquad (8)$$

$$\frac{\partial J}{\partial y_j} = \frac{\partial}{\partial y_j}\left[\frac{1}{2}\sum_{k=1}^{c}(t_k - z_k)^2\right] \qquad (9)$$

$$= -\sum_{k=1}^{c}(t_k - z_k)\frac{\partial z_k}{\partial net_k}\frac{\partial net_k}{\partial y_j} \qquad (10)$$

$$= -\sum_{k=1}^{c}(t_k - z_k) f'(net_k) w_{jk} \qquad (11)$$

Thus the sensitivity of the hidden neuron can be given as

$$\partial_j \equiv \frac{\partial J}{\partial net_j} = -\sum_{k=1}^{c}\left(\frac{\partial J}{\partial z_k}\frac{\partial z_k}{\partial net_k}\frac{\partial net_k}{\partial y_j}\frac{\partial y_j}{\partial net_j}\right) \qquad (12)$$

$$= -\sum_{k=1}^{c}\left(-\delta_k \frac{\partial net_k}{\partial y_j}\frac{\partial y_j}{\partial net_j}\right)$$

$$= f'(net_k)\sum_{k=1}^{c}(\delta_k w_{kj}) \qquad (13)$$

Finally the learning rule for the input-hidden is given as

$$\vartriangle w_{kj} = \eta x_i \delta_j = \eta x_i f'(net_j)\sum_{k=1}^{c}(w_{kj}\delta_k) \qquad (14)$$

By substituting equation (7) and (14) into equation (4), the updated weights could be obtained. The process will continue until the desired error $\theta$ goal is achieved.

$$J(W) = \frac{1}{2}\sum_{k=1}^{c}(t_k - z_k)^2 < \theta \qquad (15)$$

## 2.3 Levenberg-Marquardt BP Algorithm

The Levenberg-Marquardt back propagation algorithm has been shown to be good at training moderately sized feed forward neural networks [12]. It is an approximation of Newton's method while back propagation with gradient descent technique is a steepest descent algorithm. It updates the weights using the expression below

$$\Delta w = -\left[\mu I + \sum_{p=1}^{P} J^p(w)^T J^p(w)\right]^{-1} \nabla E(w) \qquad (16)$$

Where $J^p(w)$ is the Jacobian matrix of the error vector $e^p(w)$ evaluated in $w$, and $I$ is the identity matrix. The error vector $e^p(w)$ is the error of the network for pattern $p$, that is

$$e^p(w) = t^p - o^p(w) \qquad (17)$$

The parameter $\mu$ is increased or decreased at each step. If the error is reduced, then $\mu$ is divided by a factor $\beta$ and multiplied by $\beta$ in other case [13].

## 2.4 Genetic Algorithm

Genetic algorithms are stochastic search techniques that guide a population of solutions towards an optimum using the principles of evolution and natural genetics [7]. Genetic algorithms are inspired by the evolution of populations. In a particular environment, individuals who better fit the environment will be able to survive and hand down their chromosomes to their descendants, while less fit individuals will become extinct. The aim of genetic algorithms is to use simple representations to encode complex structures and simple operations to improve these structures. Genetic algorithms therefore are characterized by their representation and operators. In the original genetic algorithm an individual chromosome is represented by a binary string. The bits of each string are called genes and their varying values alleles. A group of individual chromosomes are called a population. [5]. Genetic algorithms are especially capable of handling problems in which the objective function is discontinuous or non differentiable, non-convex, multimodal or noisy. Since the algorithms operate on a population instead of a single point in the search space, they climb many peaks in parallel and therefore reduce the probability of finding local minima [7]. Genetic algorithm involves 3 main operators: these include selection, cross-over (recombination) and mutation.

Selection is usually the first step in a genetic algorithm process. It involves selecting chromosomes from the population to crossover and produce offspring. The selection is base on Darwin's evolution theory "survival of the fittest" that the best chromosomes should survive and produce offspring. Extracting a subset of genes from a population according a definition of quality (fitness function) is what the selection process is all about. Fitness function is the measure used to measure the optimality or closeness of the selected chromosomes to the desired ones. Many selection methods exist; they include Roulette wheel selection, Boltzmann selection, Tournament selection, rank selection, and steady state selection.

Crossover is usually the second stage. It's basically a process whereby two chromosomes combine to produce an offspring. The concept behind crossover is that an offspring may possess better qualities than its parents if the best qualities are transferred to the offspring. Crossover occurs during evolution according to a user-definable probability. Crossover selects genes from parent chromosomes and create a new offspring. Examples of crossover operators include one point crossover, two-point crossover, arithmetic and heuristic crossovers.

Mutation is an important step in genetic algorithm it helps in ensuring that the population is not stagnated at local minima. It is widely regarded as the last stage in the GA process. It helps in ensuring generic diversity from generation of population to the next. The process involves altering the gene values of one chromosome or even more. This can lead to the addition of totally new genes in the gene pools which will subsequently lead to a better result than initially obtained. Commonly used mutation operators include Flip-bit, boundary, non-uniform, uniform and Gaussian [14].

## 2.5 Optimizing BP-ANN with GA

Two factors generally influence modeling a network during the learning and training session. One is the initial interconnecting weights of the network, and another is their modified quantities [9]. Generally the initial interconnecting weights of BP ANN are often stochastically and blindly produced, this might cause the network to run into partial optimization and therefore decrease the probability to obtain the optimal solutions. Moreover, because the Delta rule is always adopted to modify the interconnecting weights of BP ANN, the convergence velocity is always slow, or sometimes the network does not even converge. These shortages of BP ANN are quite necessary to be optimized and improved [9]. The problem of partial minimum of a BP ANN can be solved by adjusting the initial interconnecting weights of the network. This can be achieved through the application of Genetic algorithm because the problem is a non linear problem. GA is a nonlinear optimization method that has very strong ability of global searching.

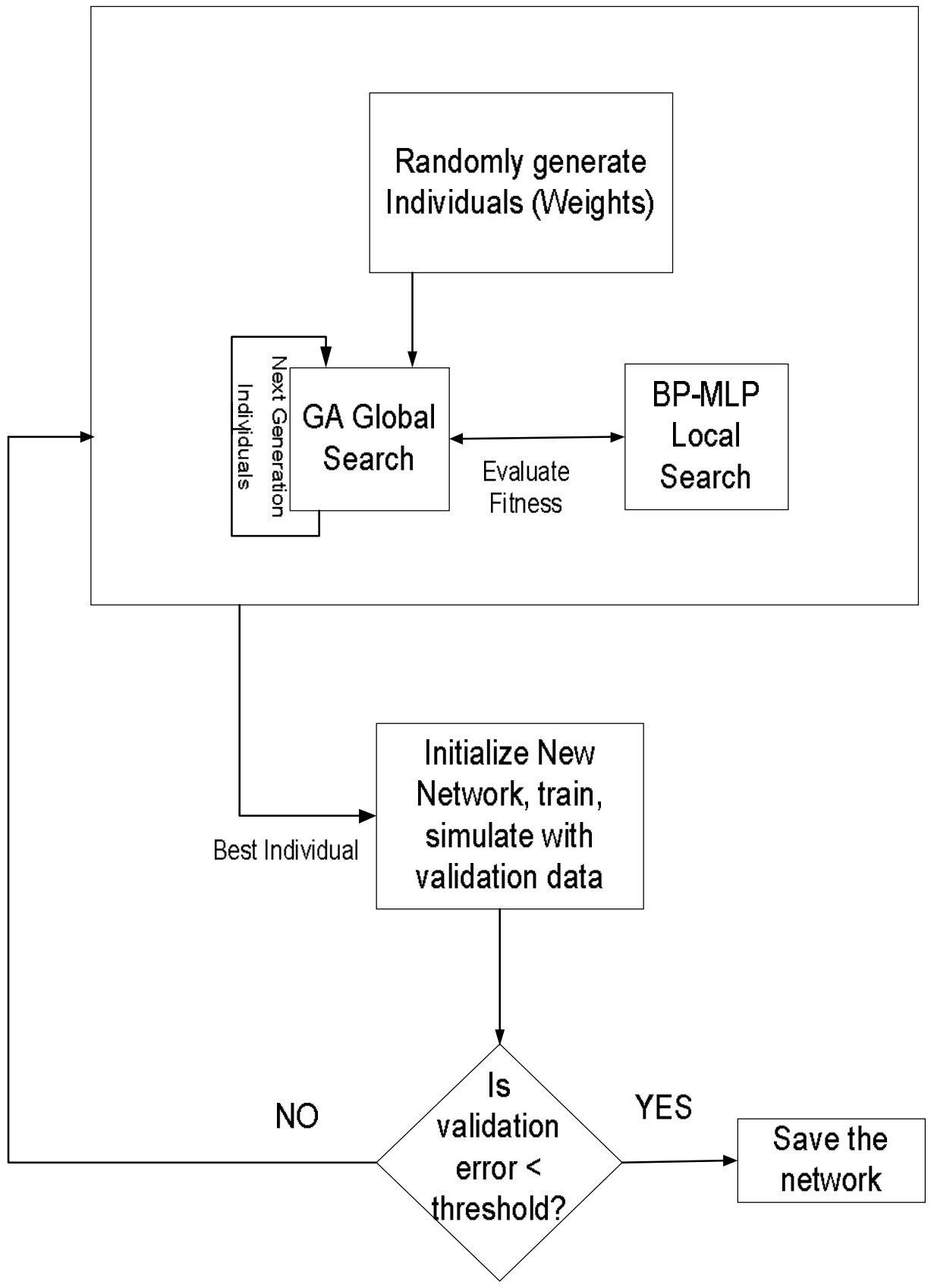

**Figure 4**: The flowchart of BP-MLP weight optimization using GA

After the structure and parameters of an ANN such as the number of layers and the number of neurons in every layer are determined, the approximate ranges of interconnecting weights and thresholds can be computed by using the algorithm of back propagation. Then, an initial population in the field of solution can be randomly produced by adopting genetic algorithm. Each group of the interconnecting weights and thresholds of BP ANN is considered to be an individual of the population and coded to be a chromosome. With the errors of BP ANN as the adaptive function, all the chromosomes are changed using GA operators (selection, crossover and mutation). The chromosomes that correspond to the best initial interconnecting weights and thresholds of BP ANN are gradually evolved. Finally, the initial interconnecting weights and thresholds optimized with GA are used into the learning and training of BP ANN again, and a better BP ANN model can be constructed. Because the initial interconnecting weights and thresholds of BP ANN are globally optimized by utilizing GA, an appropriate search space can be located in the complex solution space, and the partial minimum or non-convergence can be avoided.

## 3. MODEL IMPLEMENTATION

Based on the 12 hour spectrum occupancy measurement conducted in [10], a total of 2,700 samples were obtained since each complete sweep takes 16 seconds. These raw power level data were processed and feed to the neural network. The input and output data were randomly selected from three different services namely: GSM band, broadcasting band, and the 3G cellular band. The input is a set of power level values obtained from the spectrum analyzer during the measurement for a single channel. In [10], the channel size was given as 200 KHz. The binary hypothesis defined as $X_t = 1$ if $X_t \geq m$, otherwise $X_t = 0$, where $X_t$ are the power values from the spectrum analyzer and m is the decision threshold which was applied to get output data. The training and testing was achieved using data from the 875 MHz broadcasting band, GSM downlink 905 MHz, 3G downlink 1865 MHz and GSM uplink 890 MHz band. The entire process can be summarized in the following steps:

| **SERVICES** | **FREQUENCY** |
|---|---|
| Broadcasting | 700-806 MHz |
| GSM Uplink | 890-895 MHz |
| GSM Downlink | 905-910 MHz |
| 3G 1800 Downlink | 1865-1880 MHz |
| 3G 1900 Downlink | 1883-1890 MHz |

**Table 1**: A table of different services considered

Step 1: The time series power level data is converted into a binary time series of 0’s and 1’s using thresholding.

Step 2: The data from step one is then grouped into the different services examined. The training and testing data sets are then obtained.

Step 3: The training and testing data for the service to be predicted is selected, architecture and training parameters are also selected.

Step 4: The network is trained using MLP-BP and weight selection is optimized through the use of GA.

Step 5: The network is tested with the test data to ascertain the prediction accuracy.

The simulation was conducted using MATLAB version r2012a. After a trial and error process, the number of neurons was set at 10 which were found to be adequate. GA and MLP- BP were used during the learning process and the prediction accuracy was determined across the bands considered.

| **Parameter** | **Value** |
|---|---|
| Number of Hidden Neurons | 10 |
| Training samples | 2700 |
| Testing samples | 2700 |
| Transfer Function | Sigmoid |
| Training Algorithm | LM+GA |

**Table 2**: Parameters used in modeling

## 4. RESULTS AND DISCUSSION

In this section, the performance of the designed neural network is given. The main attraction of using neural network based spectrum prediction lies in the fact that cognitive radios can learn and train itself from historical information obtained by the cognitive radio without redesigning the whole system completely as is the case with other methods. Unlike other models such as the Markov chain approach to spectrum prediction, the neural network model need only to be updated with the most recent data as its input. This approach saves power, sensing time and manpower. Machine learning has already been proposed as an integral part of future cognitive radios, the high prediction accuracy realized in this model will greatly reduce the time required to sense whole bands. In addition, the processing power required at the base station will be also reduced. The high prediction accuracy will also drastically reduce the rate of interference between the primary and secondary users in a cognitive radio scenario. It has therefore become paramount to improve machine learning based spectrum hole prediction accuracy in order to attain high prediction accuracy. For the purpose of this work, five popular services were considered. The GSM 900 uplink channels licensed to Etisalat had a mean prediction error of about 0.035 over five channels that were selected randomly. We observed a mean prediction error of about 0.005 in the GSM 900 downlink band. 0.0179 prediction error was recorded in the 3G downlink band. 0.0004 and 0.0007 were observed in the broadcasting band and 3G downlink band (licensed to Starcomms Nigeria) respectively. The difference between the uplink and downlink prediction accuracy can be due to the fact that the base station is continuously transmitting information to the mobile users which is not the case with the uplink band. It has already been stated in [5] that the sparsely used uplink band can provide an inspiration for future deployment of some form of cognitive radio in the future. The similarity in both GSM and 3G downlink bands can be due to the similar nature of the bands in terms of their behavior. The 1900 downlink band had an error of around 0.0007 which is very low. This value might be deceiving. This band is currently being used by Starcomms Nigeria which uses a CDMA based technology for their services. The spectrum analyzer might not be able to accurately detect the presence or absence of a signal during the spectrum occupancy measurement, therefore the low value experienced might be misleading because of the low power nature of the signals in this band. In a worst case scenario whereby a signal suffers from wireless effects such shadowing, multipath, and attenuation the power considerably degrades making detection almost impossible. Overall, the performance of neural network based spectrum prediction using GA for weight optimization is good.

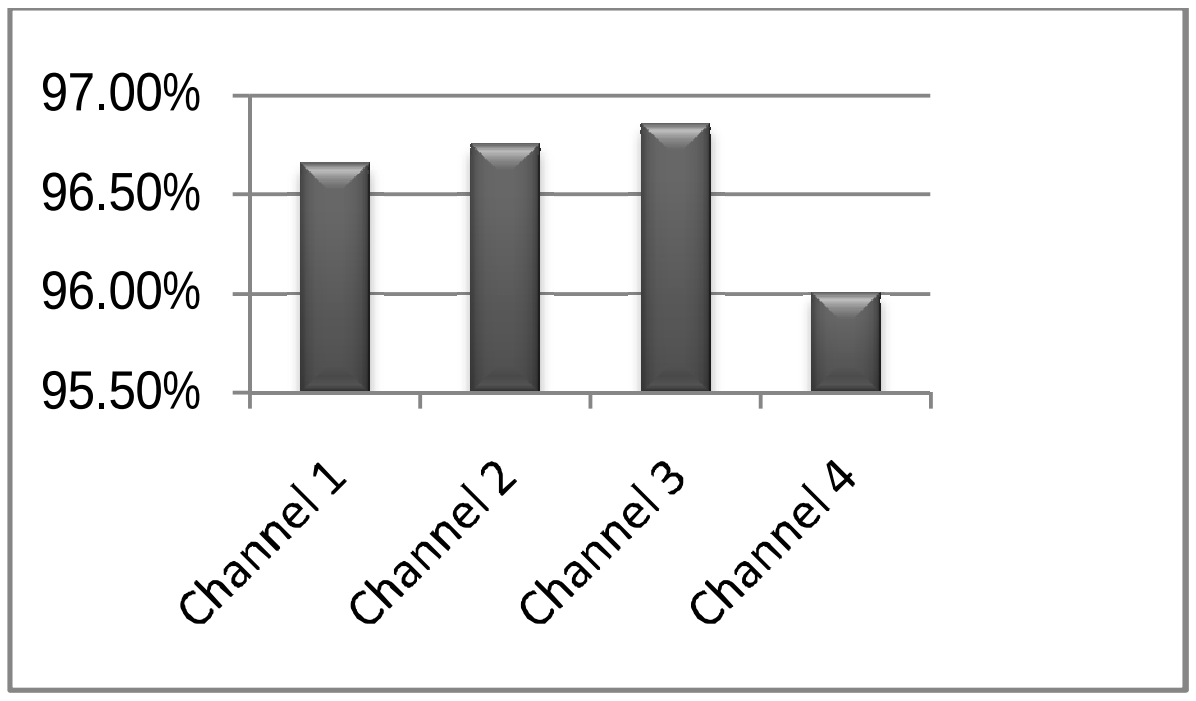

**Figure 5**: Spectrum Prediction Accuracy for 900 GSM Uplink

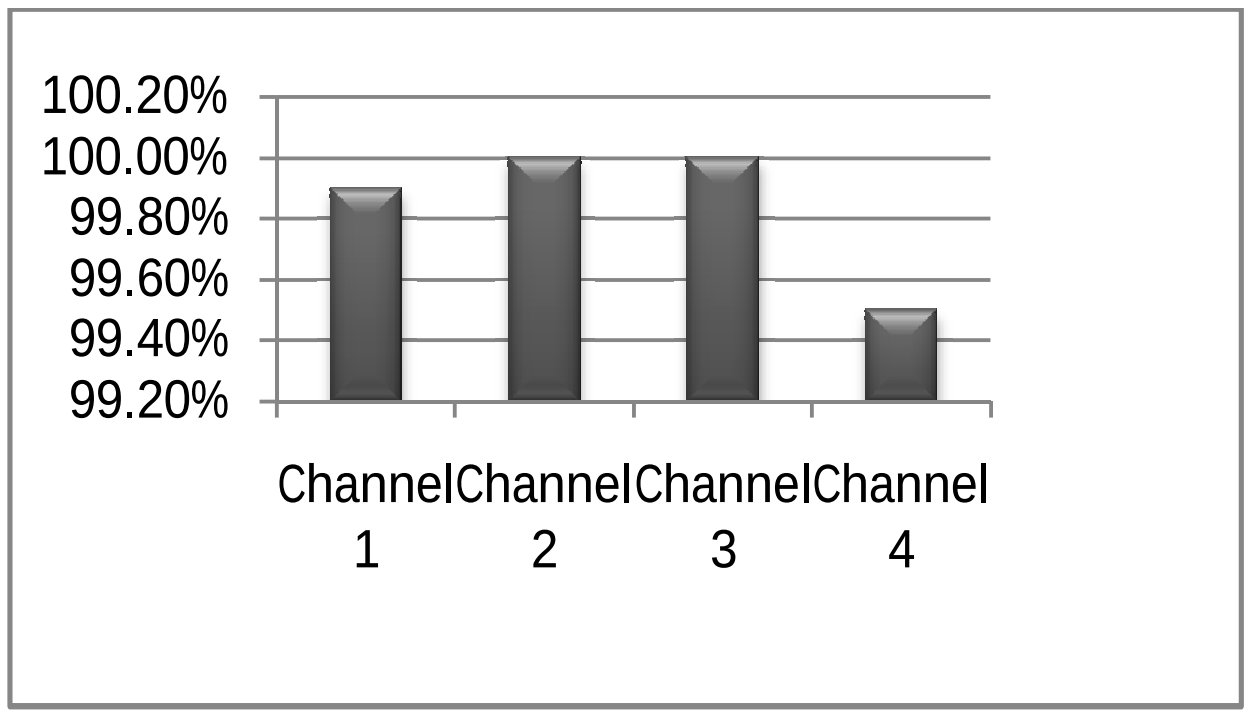

**Figure 6**: Spectrum Prediction Accuracy for 900 GSM Downlink

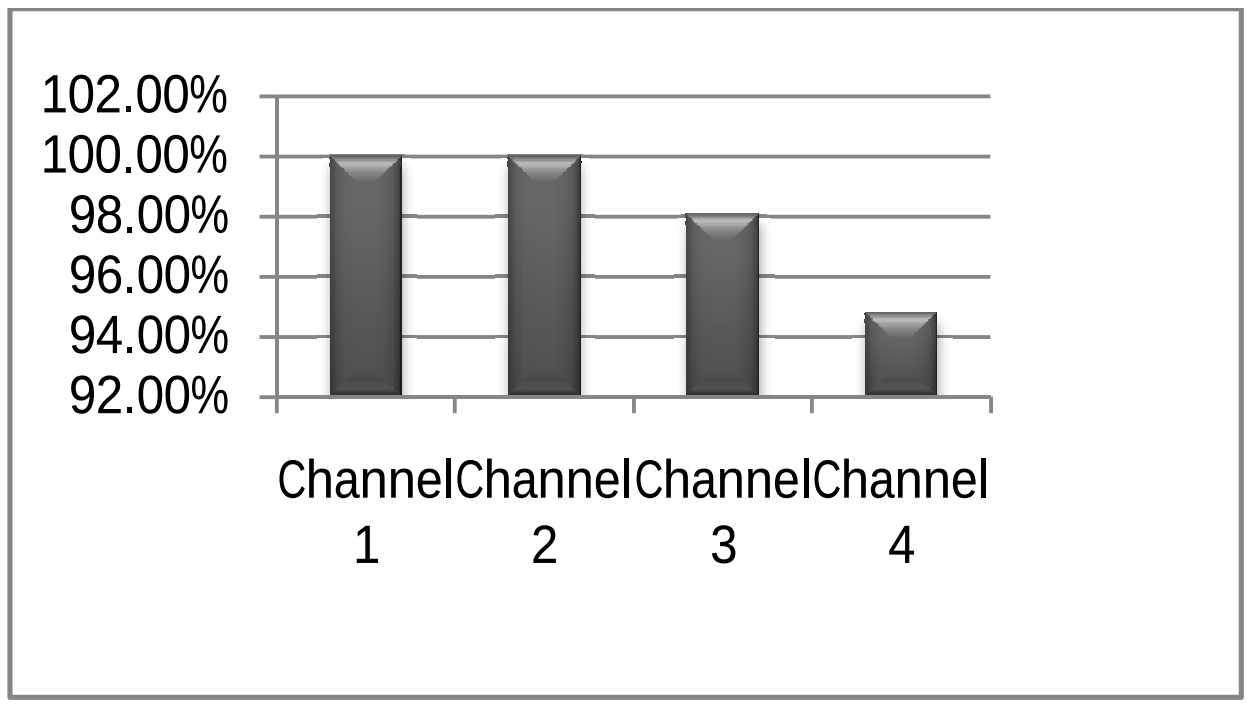

**Figure 7**: Spectrum Prediction Accuracy for 3G 1800 Downlink

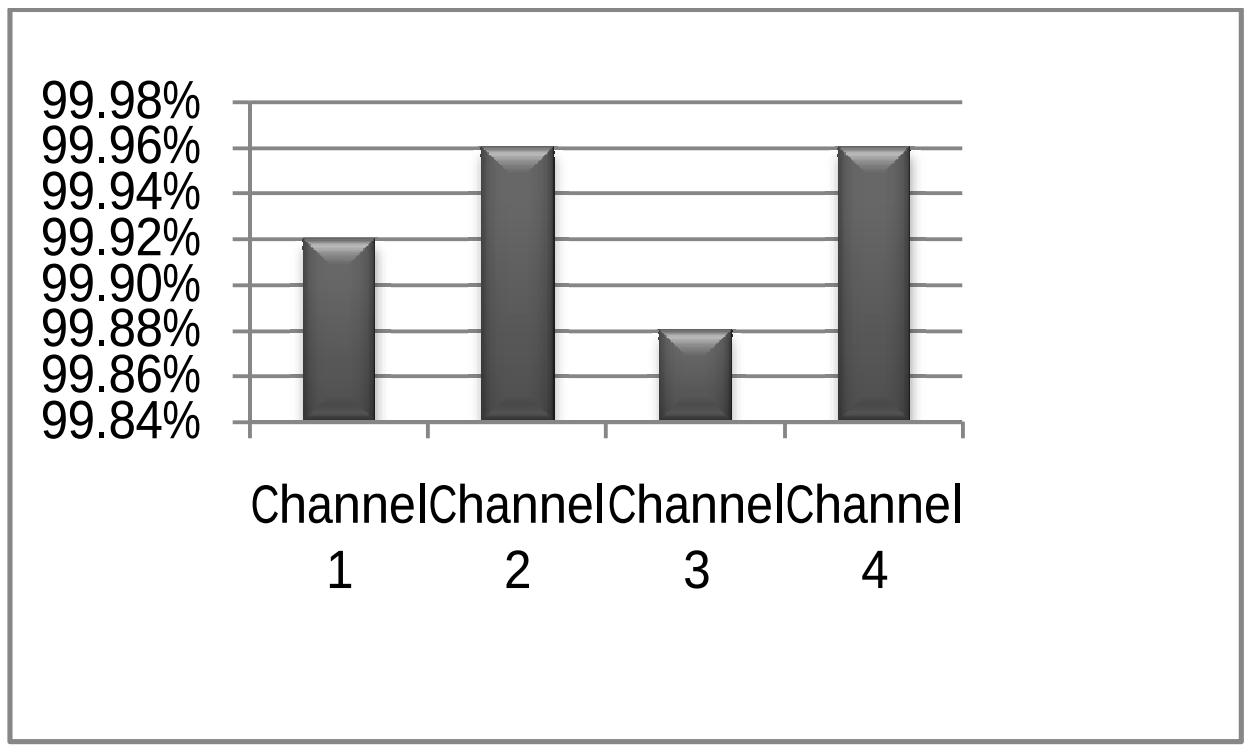

**Figure 8**: Spectrum Prediction Accuracy for broadcasting band

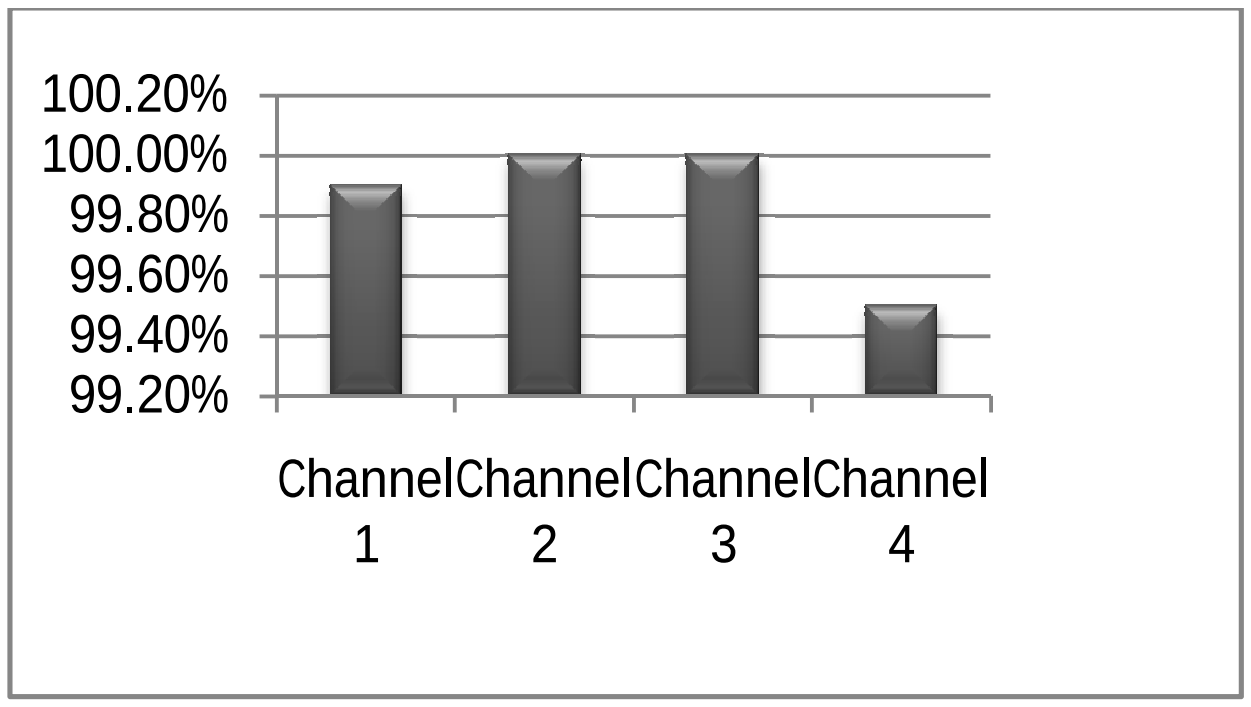

**Figure 9**: Spectrum Prediction Accuracy for 1900 3G Downlink

| SPECTRUM BAND | PREDICTION ERROR |
|---|---|
| GSM 900 UPLINK | 0.035 |
| GSM 900 DOWNLINK | 0.005 |
| 3G 1900 DOWNLINK | 0.0179 |
| BROADCASTING | 0.0004 |
| 3G 1800 DOWNLINK | 0.0007 |

**Table 1**: Prediction error for the services investigated

## 5. CONCLUSION

The importance of spectrum sensing is key to the development and eventual implementation of cognitive radios. Due to the random nature of spectrum which varies from one place to another, there is need to better understand this scarce resource so that its behavior can be predicted with little or no error. This knowledge could be obtained through extensive measurements that can help in developing models capable of predicting this behavior. Ultimately, these models could help in reducing sensing time and power consumption in cognitive radios. Unlike other models, neural network based models need not to be built from scratch once the model has been designed. Since it has already been suggested that cognitive radio will utilize a geo-location database which it will use to query information with respect to its current environment, machine learning based method could prove invaluable because of its ability to update itself whenever new information is available. Neural network based spectrum prediction has already been explored as was stated earlier; the aim of this paper was to explore the possibility of optimizing the initial weights in a neural network because they tend to run into partial optimization therefore reducing the accuracy of the deigned model. In this paper, a neural network based spectrum prediction model is presented. Unlike other models, the issue of local minimum in selecting interconnecting weights was addressed by using Genetic algorithm. Results indicate high prediction accuracy across all bands considered.

**Acknowledgement**

Our thanks go to the College of Communication Engineering, Chongqing University for their support and understanding during this work.